\documentclass[3p,12pt]{elsarticle}
\graphicspath{ {./figures/} }
\usepackage{hyperref}
\usepackage{float}
\usepackage{verbatim} 
\usepackage{apalike}
\restylefloat{figure}
\restylefloat{table}
\usepackage{bm}

\usepackage{color}

\journal{Intelligent Systems with Applications}

\biboptions{authoryear}

\begin{document}
\begin{frontmatter}

\begin{titlepage}
\begin{center}
\vspace*{1cm}

\textbf{ \large 
A Q-learning approach to the continuous control problem of robot inverted pendulum balancing}

\vspace{1.5cm}

Mohammad Safeea$^a$ (ms@uc.pt), Pedro Neto$^b$ (pedro.neto@dem.uc.pt) \\

\hspace{10pt}

\begin{flushleft}
\small  
$^a$ University of Coimbra, CEMMPRE, ARISE, Department of Mechanical Engineering, 3030-788, Coimbra, Portugal; ms@uc.pt \\
$^b$ University of Coimbra, CEMMPRE, ARISE, Department of Mechanical Engineering, 3030-788, Coimbra, Portugal; pedro.neto@dem.uc.pt \\

\vspace{1cm}
\textbf{Corresponding Author:} \\
Pedro Neto \\
University of Coimbra, CEMMPRE, ARISE, Department of Mechanical Engineering, 3030-788, Coimbra, Portugal \\
Tel: +351 239 790 767 \\
Email: pedro.neto@dem.uc.pt

\end{flushleft}        
\end{center}
\end{titlepage}

\title{A Q-learning approach to the continuous control problem of robot inverted pendulum balancing}

\author[label1]{Mohammad Safeea}
\ead{ms@uc.pt}

\author[label1]{Pedro Neto
\corref{cor1}}
\ead{pedro.neto@dem.uc.pt}

\cortext[cor1]{Corresponding author.}
\address[label1]{University of Coimbra, CEMMPRE, ARISE, Department of Mechanical Engineering, 3030-788, Coimbra, Portugal}

\begin{abstract}
This study evaluates the application of a discrete action space reinforcement learning method (Q-learning) to the continuous control problem of robot inverted pendulum balancing. To speed up the learning process and to overcome technical difficulties related to the direct learning on the real robotic system, the learning phase is performed in simulation environment. A mathematical model of the system dynamics is implemented, deduced by curve fitting on data acquired from the real system. The proposed approach demonstrated feasible, featuring its application on a real world robot that learned to balance an inverted pendulum. This study also reinforces and demonstrates the importance of an accurate representation of the physical world in simulation to achieve a more efficient implementation of reinforcement learning algorithms in real world, even when using a discrete action space algorithm to control a continuous action.
\end{abstract}


\end{frontmatter}

\section{Introduction}
\label{introduction}

Reinforcement learning (RL) is revolutionizing many fields. In robotics, RL is enabling robots to expand their autonomy and learning capabilities beyond strictly structured environments into unstructured environments \citep{sutton2018reinforcement, Ju_Juan_Gomez_Nakamura_Li_2022}.
RL promotes robot learning through trial and error, receiving feedback in the form of rewards or penalties for its actions, and learning complex actions such as human-level skills \citep{mnih2015human}, or to fly a model helicopter \citep{kim2004autonomous}. A reference study addresses policy gradient RL applied to an actuated passive dynamic walker that learned a feedback control policy for performing dynamic walking \citep{tedrake2004stochastic}. The policy was learned in twenty minutes while running online in the real robot. RL has also been successfully applied for robot learning in the context of robot shared autonomy \citep{8976128} and in position/force control of robot manipulators \citep{Perru}.

The problem of environment exploration with sparse rewards is a challenge and prevents the use of RL in some real-world tasks. Researchers propose to use demonstrations to overcome the exploration problem and learn to perform long-horizon multi-step robotics tasks \citep{8463162}. To limit the need for real world interactions while learning a policy, models are often used as simulation systems, training the policy in simulation and later transferring such policy to the real world \citep{zhao2020sim}. This process is commonly named as RL sim-to-real, featuring successfully application on robot grasping control \citep{yan2017sim} and self-supervised robotic manipulation \citep{jeong2020self}. Training and learning RL policies in simulation brings great advantages to the process, as the learning in real world involves long periods of experiments with robotic equipment to handle disturbances, requiring human supervision, besides that the learning process may damage the equipment. In summary, this process is expensive, unsafe and time consuming. On the other hand, creating accurate simulation models is challenging and sometimes even impracticable. Small errors due to under-modelling can accumulate and make the simulated robot to diverge from the real-world robot, which limits the performance when transferring the learned policies to the real system \citep{doi:10.1177/0278364913495721}. Sim-to-real learning policies often fail to transfer the learning behaviours to real world. In such a context, these systems can be improved by injecting noise in simulation and/or taking data from the real world to minimize the simulation-to-real gap and by this way improve learning policies. Recent studies demonstrate reinforcement learning solutions where only a single demonstration trajectory is required to learn behaviours \citep{pavse2020ridm}, or a new sim-to-real technique that allows end-to-end training \citep{Karnan9341149}.

The application of a discrete action space method like Q-learning in continuous state and action spaces modelling has been studied along the years \citep{8836506, gaskett1999q,Sabir_Said_Al-Mdallal_Ali_2022}. Recently, normalized advantage functions allowed to apply Q-learning with experience replay to continuous tasks, substantially improving performance on robot control \citep{gu2016continuous}. Continuous action Q-learning demonstrated good performance in heavily constrained environments when compared to RL algorithms on continuous control problems \citep{ryu2019caql}. Deep Q-network (DQN) algorithms, combining Q-learning with deep neural networks (DNNs), allows an agent to learn directly from high dimensional sensory inputs \citep{naturaa}. DQN has been used for active simultaneous localization and mapping (SLAM) for autonomous navigation featuring static and dynamic obstacles \citep{Wen2021}, motion attitude control of humanoid robots \citep{Shi2020335}, and for assembly sequence planning problem \citep{Neves_Neto_2022}. 

When solving complex high-dimensional problems, standard policy gradient methods often require many iterations and are prone to poor local optima. To solve these difficulties the algorithm Guided Policy Search (GPS) uses trajectory optimization to guide the policy away from poor local optima \citep{7138994}. In the presence of a small number of real-world samples, the resulting neural network (considering control policies represented as high-dimensional neural networks deriving robot actions based on states) is only robust in the neighbourhood of the trajectory distribution explored by real-world interactions. Generative Motor Reflexes (GMR) was introduced to tackle this exact problem, improving robustness by using motor reflexes and stabilizing actions \citep{8793775}. A robot manipulator was trained in a reaching task and in a task where the robot was expected to place a block in a hole using GMR, both in simulation and in the real-world. GMR proved itself as more robust than the GPS algorithm \citep{8793775}. As discussed above, learning real world tasks from scratch is a complex problem requiring significant training time and high sample-complexity. To tackle this issue, an asynchronous version of the Normalized Advantage Function (NAF) algorithm, parallel NAF, was used in a variety of robotic tasks both in simulation and in the real-world \citep{7989385}. This algorithm allows several robots to learn simultaneously in an asynchronous fashion, reducing the training time proportionally to the number of robots in use. RL has also been used to improve assembly efficiency recurring to dual-arm robots, reducing the efforts in robot teaching and process planning \citep{WATANABE2020107615}.

\section{Proposed approach and background}

\begin{figure}
	\begin{centering}
		\includegraphics[width=1\textwidth]{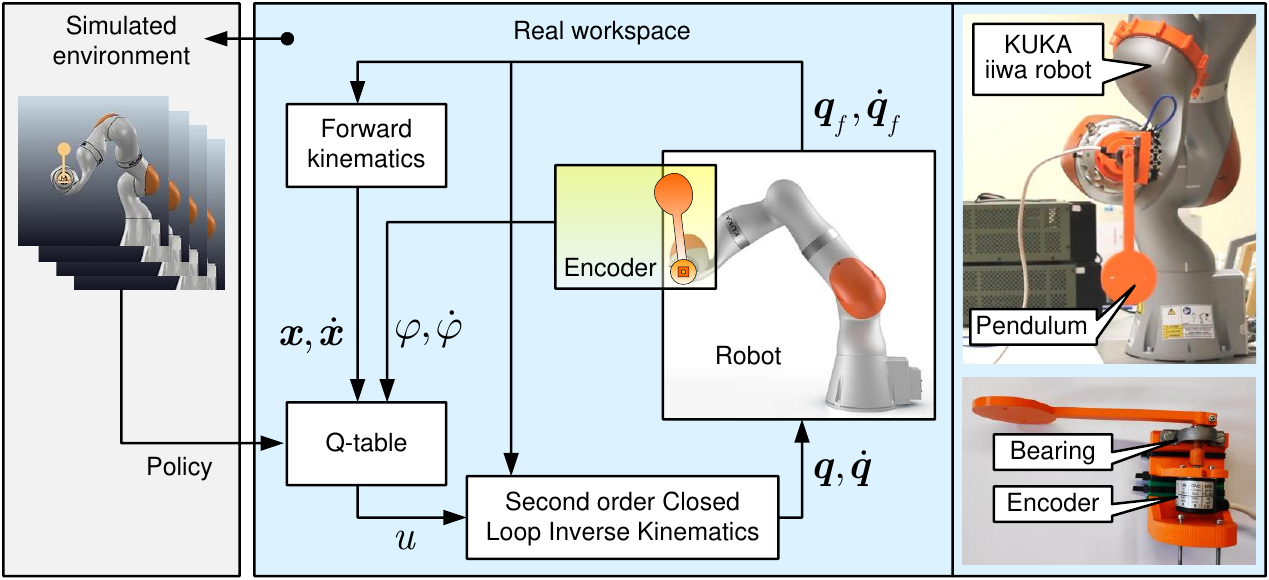}
		\par\end{centering}
	\caption{Architecture of the proposed Q-learning approach to the problem of robot inverted pendulum balancing, featuring in/out monitoring and control data. The hardware elements of the system are highlighted, namely the robot and the pendulum mechanism.\label{fig:Experimental-setup}}
\end{figure}

RL is a powerful learning tool. However, its application in real-world robotic systems is still challenging. In this study it is proposed to apply RL, tabular Q-learning, to the continuous control problem of robot inverted pendulum balancing, Fig. \ref{fig:Experimental-setup}. The proposed methodology is trained in simulation recurring to the Virtual Robot Experimentation Platform (V-REP, CoppeliaSim) \citep{rohmer2013v}, and then evaluated on a real-world robot. A mathematical model of the system dynamics is deduced by curve fitting on data acquired from the real-world system. The setup is composed by a robot manipulator and the pendulum mechanism. The pendulum and the encoder are attached to the same spindle hoisted inside the bearing. The bearing and the encoder are mounted on a 3D printed base that is attached to the flange of the robot. 

\subsection{Background}

Given a Markov Decision Process (MDP) defined by a set of states $s$,
a set of actions $a$, and a transition function $P(\grave{s}|s,a)$,
it is required to find an optimal policy $\pi^{*}$ which maximizes
the expected reward. Q-learning is a model free RL method that can be implemented to achieve the control policy $\pi^{*}$ \citep{watkins1992q}, while the agent learns during its interaction with the environment.
The table covers the discretized state-action space, containing scalar values indicating how fit an action is for a given state. During the learning process the table is updated in each iteration to maximize the expected reward while the agent is interacting with the environment. The learning process is represented mathematically by:

\begin{equation}
Q\left(s,a\right)=\left(1-\alpha\right)Q\left(s,a\right)+\alpha\left(r+\gamma max_{\grave{a}}\left(Q\left(\grave{s},\grave{a}\right)\right)\right)
\end{equation}

\noindent Where $Q\left(s,a\right)$ is the state-action value for
the action $a$ at the state $s$, $\alpha$ is the learning rate,
$r$ is the reward, $\gamma$ is the discount ratio, and $\grave{s}$
is the resulting (observed state) after applying the action $a$.
The algorithm can be applied for controlling dynamical systems where
the resulting policy can be used for generating the control signals
required to stabilize the physical system.

\section{Methodologies}
\subsection{Robot acceleration tracking \label{sec:Acceleration-tracking-with}}

The accelerations required to balance the pendulum are the RL actions. However, the robot is a series of rigid links connected by joints, and thus, the robot moves by actuating such joints. The robot can track the commanded acceleration of the flange by using a second order Closed Loop Inverse Kinematics (CLIK) algorithm \citep{siciliano1990closed}. From differential kinematics the following relationship between the acceleration of the joints and the acceleration of the end-effector (EEF) can be established:

\begin{equation}
\Gamma=\boldsymbol{\mathrm{J}}\ddot{\boldsymbol{q}}+\dot{\boldsymbol{\mathrm{J}}}\dot{\boldsymbol{q}}\label{eq:accel}
\end{equation}

\noindent Where, $\Gamma$ is the acceleration of the EEF (robot flange), $\boldsymbol{\mathrm{J}}$ is the Jacobian matrix of
the robotic manipulator, $\ddot{\boldsymbol{q}}$ is the angular acceleration
vector of the robot joints, $\dot{\boldsymbol{\mathrm{J}}}$ is the
time derivative of the Jacobian matrix, and $\dot{\boldsymbol{q}}$
is the angular velocity vector of the robot joints. The robot
is 7  degrees of freedom (DOF) redundant manipulator. However, to simplify the experiment we fix the redundant joint (third) of the robot. Thus, the resulting Jacobian is square and the joints angular accelerations are calculated:

\begin{equation}
\ddot{\boldsymbol{q}}=\boldsymbol{\mathrm{J}}^{-1}(\Gamma+\boldsymbol{\mathrm{K}}_{d}\dot{\boldsymbol{e}}+\boldsymbol{\mathrm{K}}_{p}\boldsymbol{e}-\dot{\boldsymbol{\mathrm{J}}}\dot{\boldsymbol{q}})\label{eq:CLICK}
\end{equation}

\noindent Where $\boldsymbol{\mathrm{K}}_{d}$ and $\boldsymbol{\mathrm{K}}_{p}$
are the derivative and proportional gains, both positive definite matrices.
The position/orientation error at the EEF level is $\boldsymbol{e}$, and $\dot{\boldsymbol{e}}$ is the linear/angular velocity error at the
EEF.

\subsection{Mathematical model of the pendulum\label{sec:Mathematical-model-of}}

The state of the pendulum at any instant is specified by the vector $\boldsymbol{s}$:

\begin{equation}
\boldsymbol{s}=\left[\begin{array}{c}
x\\
\dot{x}\\
\varphi\\
\dot{\varphi}
\end{array}\right]
\end{equation}

\noindent Where $x$ and $\dot{x}$ are the linear displacement and velocity of the robot's flange along the $x$ axis of its base, respectively. The angular position $\varphi$ and velocity $\dot{\varphi}$ of the pendulum are measured from the vertical upright position (in the unstable fixed point). Thus, the control problem requires to stabilise the system at the state $\boldsymbol{s}^{*}=[0,0,0,0]^{\mathrm{T}}$. On the other hand, the robot is controlled at the joints position level. By moving its joints, the robot constrains the motion of its flange along the $x$ axis and tries to balance the pendulum in the unstable fixed point (vertical upright position). Thus, the control input $u$ is the acceleration of the flange along the $x$ axis:

\begin{equation}
u=\ddot{x}
\end{equation}

The tracking acceleration of the robot along the $x$ axis is given by:

\begin{equation}
\grave{u}=n^{\mathrm{T}}\grave{\Gamma}
\end{equation}

\noindent Where $n^{\mathrm{T}}$ is equal to the vector $[1,0,0,0,0,0]$,
and $\grave{\Gamma}$ is the actual (tracking) acceleration at the
robot flange due to joints acceleration/velocities (\ref{eq:accel}),
as a result of the control command $u$, and by considering the error
dynamics $e$ and $\dot{e}$. In such a case the state transition equation
of the proposed system near the unstable fixed point is given by:

\begin{equation}
\dot{\boldsymbol{s}}=\left[\begin{array}{cccc}
0 & 1 & 0 & 0\\
0 & 0 & 0 & 0\\
0 & 0 & 0 & 1\\
0 & 0 & \frac{mgl}{I} & \frac{-b}{I}
\end{array}\right]\boldsymbol{s}+\left[\begin{array}{c}
0\\
1\\
0\\
\frac{-ml}{I}
\end{array}\right]\grave{u}\label{eq:state-equation}
\end{equation}

\noindent Where $I$ is the moment of inertia of the pendulum and
the rotary part of the encoder around the rotation axis, $b$ is the
viscus friction coefficient, $m$ is the mass of the pendulum, $g$
is the gravity acceleration, $l$ is the distance from the center
of mass of the pendulum to the rotation axis, and $\grave{u}$ is the actual (tracking)
acceleration of the robot's flange along the $x$ axis.

\begin{figure}
	\begin{centering}
		\includegraphics[width=0.6\textwidth]{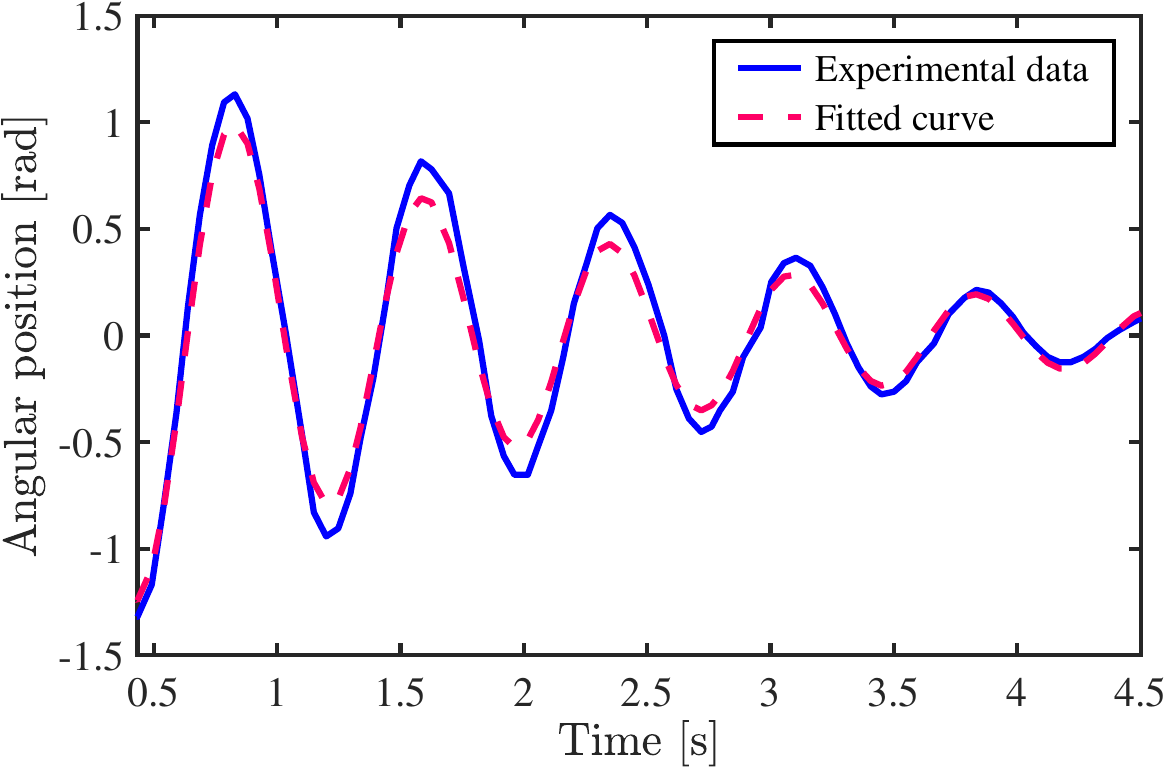}
		\par\end{centering}
	\caption{Free oscillation of the pendulum. Encoder experimental data are plotted in continuous line, while the fitted curve is represented in dashed line.\label{fig:Free-oscillation-curve}}
\end{figure}

For the proposed system, the state equation (\ref{eq:state-equation}) is used, so that the system learns the commanded accelerations $u$ required to balance the pendulum according to both the pendulum's dynamics and the robot's acceleration tracking dynamics. However, from (\ref{eq:state-equation}) it is noticed that there are parameters that need to be estimated, specifically $I$ and $b$. In such a context, it was conducted an experiment to achieve a precise estimation of the aforementioned constants. The mathematical equation of a pendulum rotating around a fixed axis without an external excitation is given by:

\begin{equation}
I\ddot{\theta}+b\dot{\theta}+mglsin(\theta)=0\label{eq:damped_motion_equation_of_a_pendulum} \label{eq:state-equation2}
\end{equation}

\noindent  Where $\theta$ is the pendulum's angle measured from the vertical down position (the stable fixed point), so that $\theta+\varphi=\pi$. The angular velocity and acceleration of the pendulum under free oscillation are represented by $\dot{\theta}$ and $\ddot{\theta}$, respectively. From equation (\ref{eq:state-equation2}), the values of $I$ and $b$ can be estimated by the angular coordinate $\theta$ of the pendulum (from the encoder) while it is freely oscillating due to an initial angular displacement, i.e., inducing free oscillations in the pendulum while capturing $\theta$, and then by fitting a curve into the resulting data according to the mathematical model in (\ref{eq:damped_motion_equation_of_a_pendulum}). Fig. \ref{fig:Free-oscillation-curve} shows a plot of the captured data in addition to the resulting fitted curve, corresponding to the resulting estimates of $I$ and $b$. In this experiment, the pendulum has a small distance from the centre of mass to the rotation axis $l$. This is reflected on the characteristics of the plot in Fig. \ref{fig:Free-oscillation-curve}, where the period of the natural oscillation is around 0.75 seconds, and therefore, the task of balancing the inverted pendulum is challenging, even for a human.

\section{Results and discussion}

\begin{figure*}
	\includegraphics[width=1\textwidth]{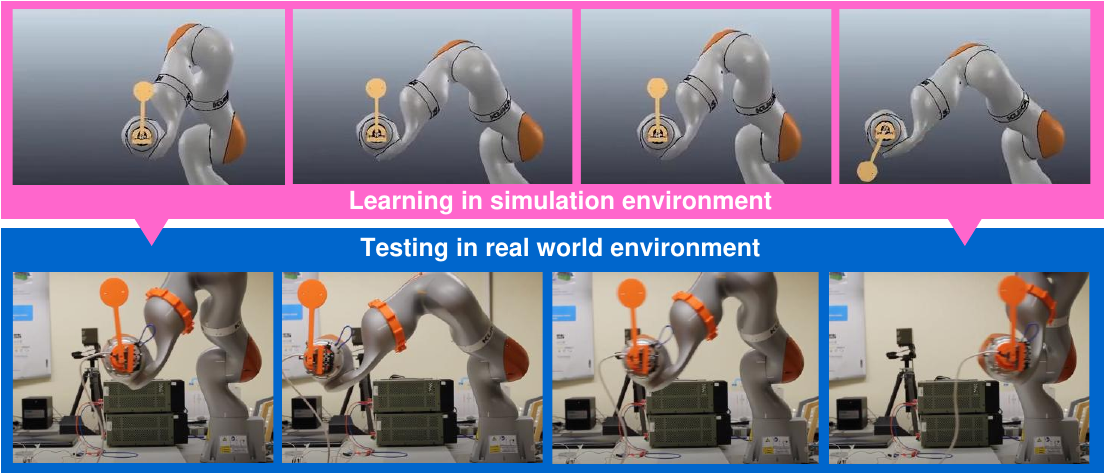}
	
	\caption{RL policy training in V-REP simulation environment and policy evaluation on the real world environment.\label{fig:simulation_and_real}}
\end{figure*}

\subsection{Discretization}

Fig. \ref{fig:simulation_and_real} shows the policy training process in simulation and the robotic system operating in real world environment (video in multimedia materials). Different quantities were discretized, namely:
\begin{enumerate}
\item The control command $u$ (linear acceleration of the robot flange) was discretized into eight different actions: $\left\{ -2, -1.5, -1, -0.5, 0.5, 1, 1.5, 2\right\} $ $m/s^{2}$; 
\item The state of the system, i.e., the angular position of the pendulum $\varphi$ (taken from the vertical upright position) was discretized into six different states in the intervals: $]-11{^\circ},-5{^\circ}[$, $[-5{^\circ},-1{^\circ}[$, $[-1{^\circ},0{^\circ}[$, $[0{^\circ},1{^\circ}[$, $[1{^\circ},5{^\circ}[$ and $[5{^\circ},11{^\circ}[$ degrees;
\item The angular velocity of the pendulum $\dot{\varphi}$ was discretized into five different states in the intervals: $]-\infty,-50]$, $]-50,-10]$, $]-10,10[$, $[10,50[$ and $[50,+\infty[$ degrees per second;
\item The linear position of the flange $x$ was discretized into three different states in the intervals: $]-0.22,-0.08]$, $]-0.08,0.08[$ and $[0.08,0.22[$ meters;
\item The linear velocity of the flange $\dot{x}$ was discretized into three different states in the intervals: $]-\infty,-0.5]$, $]-0.5,0.5[$ and $[0.5,+\infty[$ $m/sec$. 
\end{enumerate}
A failure is declared when the angular position $\varphi$ exceeds any of the limits $\left\{ +11{^\circ},-11{^\circ}\right\} $ degrees, or when the flange exceeds any of the limits $\left\{ +0.22,-0.22\right\} $ meters.

\subsection{Simulation}

The simulations were performed in V-REP, Fig. \ref{fig:simulation_and_real}, using the remote-API to control it from MATLAB where the mathematical model of the system was implemented, equations (\ref{eq:state-equation}) and (\ref{eq:CLICK}). The Q-learning control policy was trained in 10,000 episodes, where for each episode was injected a small noise into the estimated parameters $I$ and $b$. A time step $h=0.01$ seconds is used and a small noise is injected into the error $e$ for each iteration.

\subsection{Real World System}

\begin{figure}
	\begin{centering}
		\includegraphics[width=0.58\textwidth]{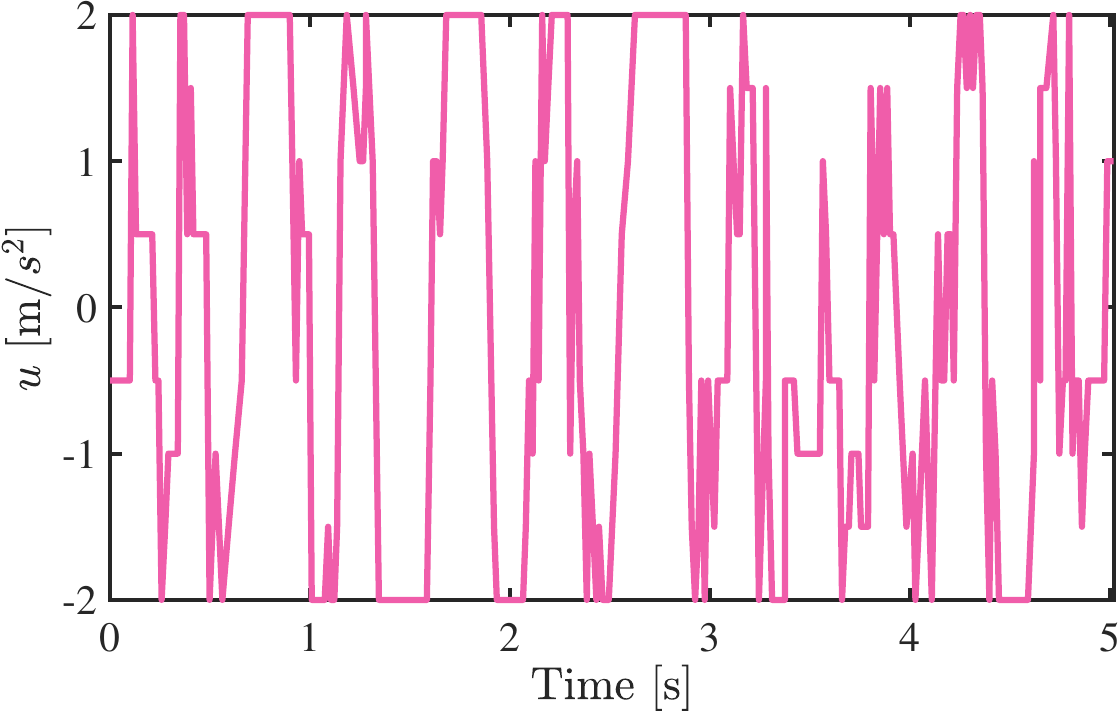}
		\par\end{centering}
	\caption{Discretized control commands $u$ featuring the linear acceleration of the robot flange.\label{fig:Control-command-in}}
\end{figure}

\begin{figure}
	\begin{centering}
		\includegraphics[width=0.58\textwidth]{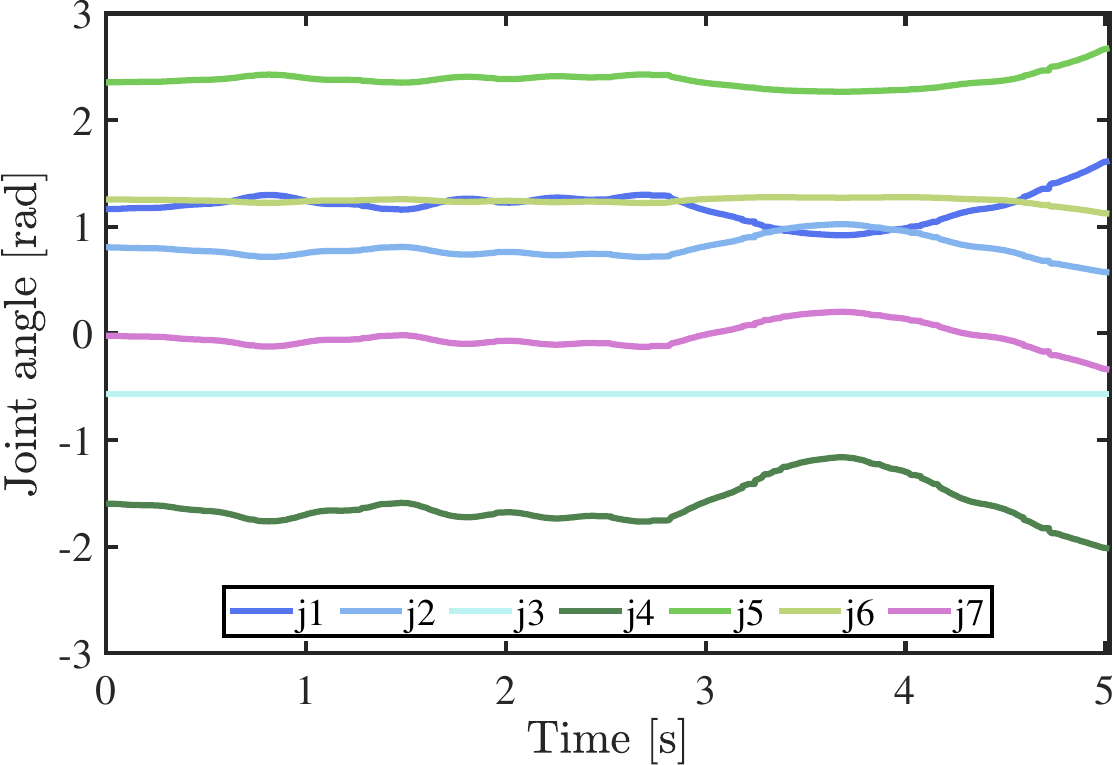}
		\par\end{centering}
	\caption{Robot joints positions reflecting the control commands in Fig. \ref{fig:Control-command-in} \label{fig:RobotJointsPositionsPlot}.}
\end{figure}

The policy trained in simulation is transferred to control the real system. A computer running a real time operating system is used to control the robot \citep{8542757}, and a trajectory generation algorithm is implemented on the robot controller. The computer is also used to acquire data from the encoder and the motion feedback from the robot.
From the acquired data, an appropriate action $u$ is chosen according to the policy trained in simulation. Afterwards, the robot angular quantities, accelerations, velocities and positions are
calculated according to the CLIK algorithm. The robot control commands defined are shown in Fig. \ref{fig:Control-command-in} and the resulting robot commanded angular positions (robot joints) are in Fig. \ref{fig:RobotJointsPositionsPlot}. 

\begin{figure}
	\begin{centering}
		\includegraphics[width=0.65\textwidth]{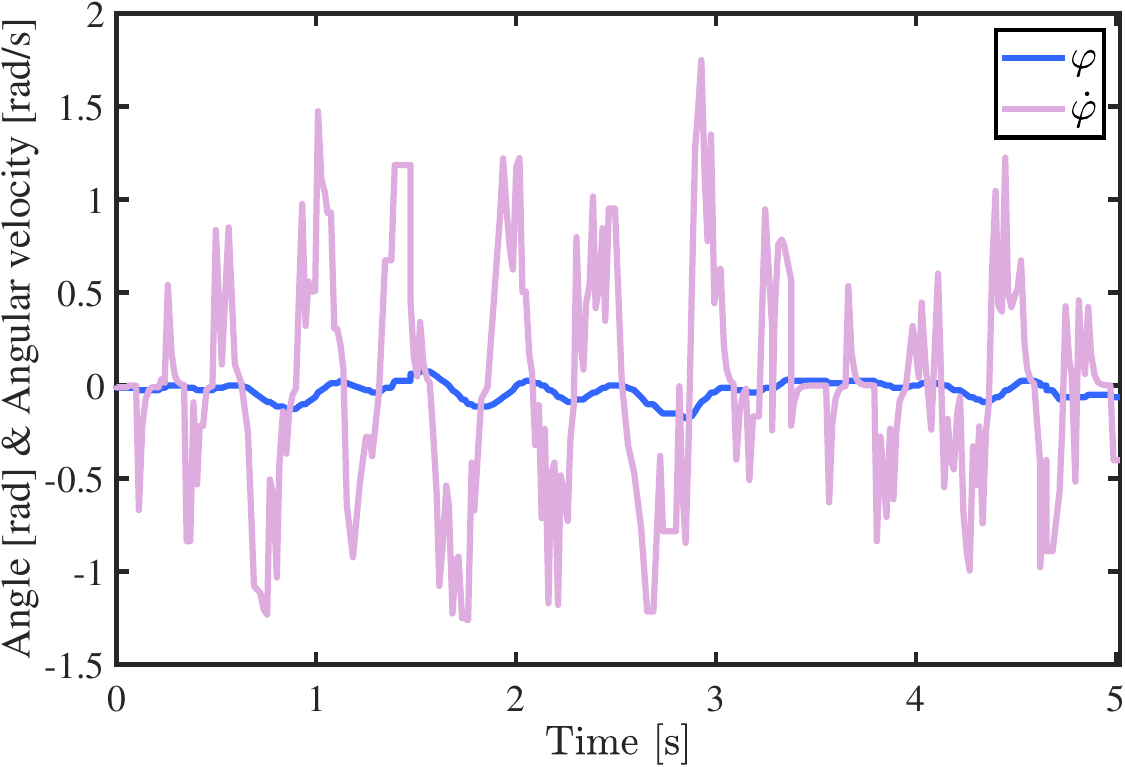}
		\par\end{centering}
	\caption{Angular position and velocity of the pendulum in real world experiments. \label{fig:The-angular-position}}
\end{figure}

\begin{figure}
	\begin{centering}
		\includegraphics[width=0.65\textwidth]{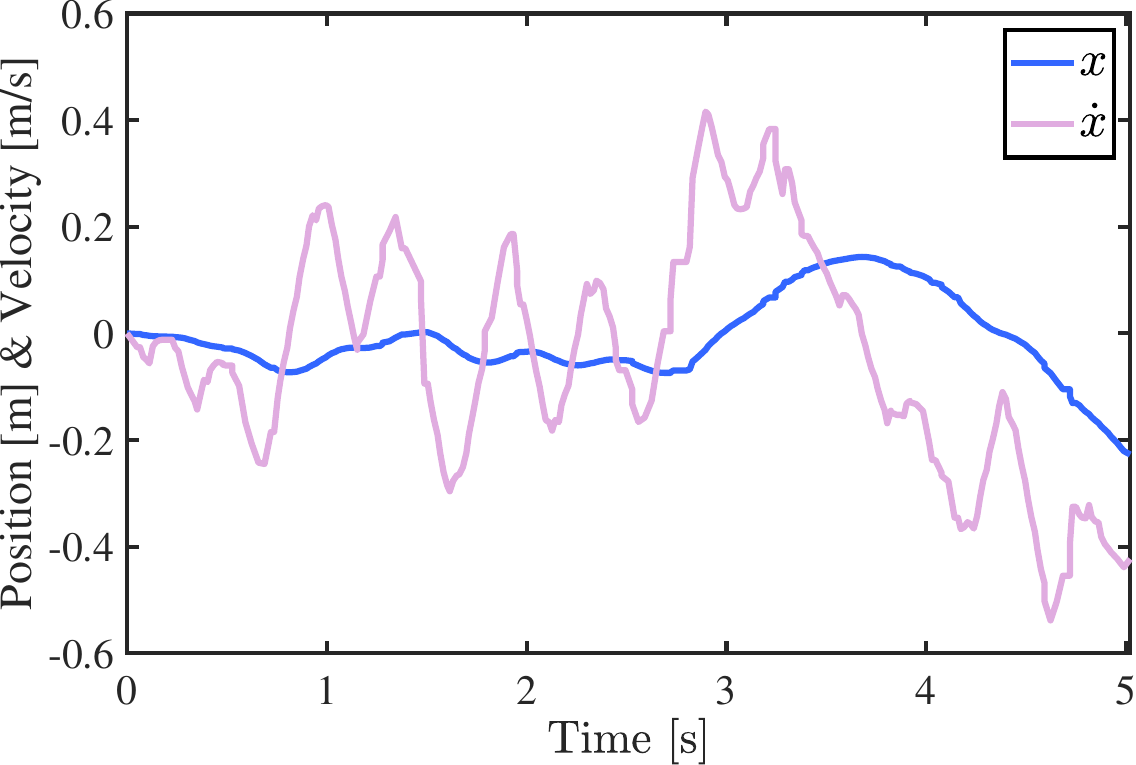}
		\par\end{centering}
	\caption{Linear displacement and velocity of the flange along $x$
		axis in real world experiment\label{fig:The-linear-position}.}
\end{figure}

The angular position/velocity of the pendulum is shown in Fig. \ref{fig:The-angular-position}. From this figure it can be seen that the angular position of the pendulum stays inside the limits. The linear position/velocity plot of the robot's flange is shown in Fig. \ref{fig:The-linear-position}, where the linear position is calculated as the displacement from a starting home position. In the proposed experiment the system was able to balance the pendulum for the first 5 seconds, however, it failed eventually, due to the violation of the flange's position, exceeding the limit at $-0.22$ meters, Fig. \ref{fig:The-linear-position} (at this instant the pendulum's angle is $-3.44$).

\subsection{Discussion}

The proposed approach, training the control policy using Q-learning in simulation environment and transferring it to real world, offers various advantages, namely:
\begin{enumerate}
	\item Speeding up the training process (time efficiency);
	\item Reducing the risk to damage hardware;
	\item The ability to start the process from various initial states which might not be easily performed using real equipment.
\end{enumerate}
 
The RL policy was trained using 10,000 episodes. While it took few minutes for training on a computer, performing the same number of trials on the real robotic system would take considerable amount of time. Note that the required time for the experiment also includes the preparation phase, to be performed at the beginning of each trial (training episode). For example, the encoder used in the experiment is an incremental encoder, consequently, at the beginning of each training episode it shall be referenced by allowing the pendulum to settle in the vertical downward position (stable fixed point) which is used as the reference. The referencing is done by pressing a button. The server application on the robot should be running, and the control program on the computer turned on. Finally, a time-delay is used in the control program to allow the user to place the pendulum in the vertical upright position before starting the control loop.

The real world system requires the user to set the pendulum upwards near the vertical by hand at the beginning of each training episode. Afterwards, the user shall release the pendulum immediately when the control loop is activated (prompted by a message on the computer screen). This procedure, the coordination between the sight of the visual notification and the immediate release, proved challenging for the users, especially because in this particular application the period of the natural oscillations of the pendulum is small, resulting in the pendulum falling rapidly and colliding with the user's hand, ruining the experiment.

An accurate dynamical model is key to achieve good results in sim-to-real transfer. However, the process to obtain such model is challenging, as well as the process to achieve good results in the presence of an inaccurate model. It is shown that even tabular Q-learning can achieve a feasible solution despite the loss of information due to discretization.

\section{Conclusion}
This study evaluated the performance of a discrete action space RL method (Q-learning) to the continuous control problem of robot inverted pendulum balancing. The control policy, trained in simulation environment and transferred to the real robot, demonstrated feasible when the system is accurately modelled. In such a condition, it can be concluded that a discrete action space algorithm can be used to control a continuous action. The advantages the simulation brings to the process are multiple, namely the reduced training time, reduced risk to damage hardware and the ability to start the training from various initial states. The learned policy can fail in some conditions due to some remaining unmodeled dynamics and the perturbations in the real system which will always exist. Future work will be dedicated to improving the control policy with more information from the real world environment.

\section*{Acknowledgements}
This research is sponsored by national funds through Fundação para a Ciência e a Tecnologia, under the project UIDB/00285/2020 and LA/P/0112/2020.

\bibliography{RLfirst}

\end{document}